\documentclass{article}

\usepackage{PRIMEarxiv}

\usepackage[utf8]{inputenc} % allow utf-8 input
\usepackage[T1]{fontenc}    % use 8-bit T1 fonts
\usepackage{cite}
\usepackage{amsmath,amssymb,amsfonts}
\usepackage{algorithmic}
\usepackage{graphicx}
\usepackage{textcomp}
\usepackage{xcolor}
\usepackage{array}
\usepackage{multirow}
\usepackage{hyperref}
\def\BibTeX{{\rm B\kern-.05em{\sc i\kern-.025em b}\kern-.08em
    T\kern-.1667em\lower.7ex\hbox{E}\kern-.125emX}}

\usepackage{cite}
\usepackage{balance}
\usepackage{authblk}
\usepackage{booktabs}
\usepackage{algorithmic}
\usepackage{graphicx}
\usepackage{textcomp}
\usepackage{xcolor}
\usepackage{url}
\usepackage{graphicx}
\usepackage{subcaption}
\usepackage{amsmath}
\usepackage{enumitem}
\usepackage{graphicx}
\usepackage{textcomp}
\usepackage{xcolor,colortbl}
\usepackage{verbatim}

\definecolor{LigthGray}{gray}{0.95}
\definecolor{LighterGreen}{rgb}{0.9, 1, 0.9}
\definecolor{LightGreen}{rgb}{0.6, 1, 0.6}
\definecolor{LighterRed}{rgb}{1, 0.9, 0.9}
\definecolor{LightRed}{rgb}{1, 0.6, 0.6}

\definecolor{GRU-FCN}{rgb}{1, 0.8, 0.8}
\definecolor{RCLSTM}{rgb}{0.8, 1, 0.8}
\definecolor{LSTM-FCN}{rgb}{0.8, 0.8, 1}
\definecolor{ETSformer}{rgb}{1, 0.85, 0.75}
\definecolor{Mean}{rgb}{0.9, 0.7, 1}
\definecolor{LSTM}{rgb}{0.7, 1, 0.89}
\definecolor{GRU}{rgb}{0.9, 1, 0.7}

\newcolumntype{C}[1]{>{\centering\arraybackslash}p{#1}}
\newcolumntype{L}[1]{>{\raggedright\arraybackslash}p{#1}}

\usepackage{todonotes}

\graphicspath{{media/}}     % organize your images and other figures under media/ folder

\title{ Comparative Analysis of Deep Learning Models for Real-World ISP Network Traffic Forecasting  \\
\thanks{This work was supported by the Student Summer Research Program 2024 of FIT CTU in Prague. Furthermore, this research was partially funded by the Ministry of Interior of the Czech Republic in project ``Flow-Based Encrypted Traffic Analysis'' (VJ02010024), by the Ministry of Education, Youth and Sports of the Czech Republic in project ``e-Infrastructure CZ'' (LM2023054), and also by the Grant Agency of the CTU in Prague, grant No. SGS23/207/OHK3/3T/18 funded by the MEYS of the Czech Republic. Computational resources were provided by the e-INFRA CZ project (ID:90254), supported by the Ministry of Education, Youth and Sports of the Czech Republic.}
}

\author[1,3]{Josef Koumar}
\author[1]{Timotej Smoleň}
\author[2]{Kamil Jeřábek}
\author[1,3]{Tomáš Čejka}
\affil[1]{Czech Technical University in Prague, Faculty of Information Technology, Czech Republic \authorcr Corresponding author email: {\tt koumajos@fit.cvut.cz}}
\affil[2]{Brno University of Technology, Faculty of Information Technology, Czech Republic}

\affil[3]{CESNET, a.l.e., Prague, Czech Republic}

\begin{document}
\maketitle

\begin{abstract}
  
Accurate network traffic forecasting is essential for Internet Service Providers (ISP) to optimize resources, enhance user experience, and mitigate anomalies. This study evaluates state-of-the-art deep learning models on CESNET-TimeSeries24, a recently published, comprehensive real-world network traffic dataset from the ISP network CESNET3 spanning multivariate time series over 40 weeks. Our findings highlight the balance between prediction accuracy and computational efficiency across different levels of network granularity. Additionally, this work establishes a reproducible methodology that facilitates direct comparison of existing approaches, explores their strengths and weaknesses, and provides a benchmark for future studies using this dataset.

\end{abstract}

% keywords can be removed
\keywords{neural networks \and deep learning \and network traffic forecasting \and network traffic prediction \and network monitoring}

\section{Introduction}

Traffic monitoring is a cornerstone of effective network management and cybersecurity, providing Internet Service Providers (ISPs) with critical insights to detect anomalies, mitigate congestion, and maintain network performance \cite{d2019survey}. The surge in video streaming, cloud computing, and online gaming is driving rapid growth in internet usage, contributing to increasingly complex and less predictable network traffic. Efficient network monitoring allows ISPs to maintain service quality, mitigate security risks, and optimize bandwidth in real time \cite{hassan2024optimizing}.

However, real-time monitoring alone is insufficient for proactively managing network resources. To anticipate variations in demand and prevent service disruptions, ISPs increasingly adopt advanced forecasting techniques to predict traffic patterns and optimize resource allocation in advance \cite{oliveira2020adaptive}.

Accurate traffic forecasting allows ISPs to efficiently allocate resources, scale network capacity, and sustain service quality under fluctuating loads \cite{oliveira2020adaptive}. The rise of diverse, high-bandwidth services has significantly increased network traffic variability. Traditional models like ARIMA and exponential smoothing, which assume linearity, struggle with ISP data due to prevalent non-linear and high-frequency fluctuations, especially during peak traffic hours \cite{ahmed2016survey}.

These limitations have driven the adoption of deep learning models, particularly neural networks, which excel at capturing complex temporal dependencies across various forecasting domains~\cite{wu2021current}. By utilizing deep learning’s ability to model complex patterns, ISPs can gain deeper insights into usage trends and anomalies, significantly enhancing traffic forecasting capabilities.

Wu et al.~\cite{wu2021current} attribute this development to the increasing adoption of deep learning models. However, assessing recent improvements in forecasting methods for network traffic monitoring remains challenging due to the lack of long-term real-world datasets. In their survey, Ferreira et al.~\cite{ferreira2023forecasting} identify the absence of a reference dataset as a key obstacle to performance evaluation. As a result, most approaches are evaluated on publicly available datasets that are artificially generated. Wu et al.~\cite{wu2021current} demonstrate that evaluating novel forecasting approaches on synthetic datasets can create the illusion of nonexisting progress.

To address these challenges, this study presents a comparative evaluation of advanced deep learning models using a public dataset, CESNET-TimeSeries24 \cite{koumar2024cesnettimeseries24timeseriesdataset}, created from real-world ISP traffic. In this comparative study, we focus on answering the following four research questions:

\textbf{RQ1:} What is the performance of state-of-the-art deep learning models for real-world network traffic forecasting?

\textbf{RQ2:} How do the training and prediction windows impact the forecasting performance of state-of-the-art deep learning models?

\textbf{RQ3:} How does the granularity of aggregation (e.g., individual IP addresses, subnets, or institutions) impact the forecasting performance of state-of-the-art deep learning models?

\textbf{RQ4:} How do different monitored metrics (e.g. transmitted data, TCP / UDP ratio, average duration) affect the forecasting performance of state-of-the-art deep learning models?

The dataset's scale and diversity enable a thorough assessment across multiple time series metrics and hierarchical network levels. By combining rigorous methodological practices with open-access resources, this work establishes fair, transparent, and reproducible benchmark for network traffic forecasting models. The purpose of this work is to encourage the adoption of standardized datasets in future research, ensuring fair comparisons and practical insights. Moreover, to facilitate reproducibility, we publish source codes of models and experiments on GitHub\footnote{\url{https://github.com/koumajos/isp-forecasting-benchmark}}.

The rest of this paper is structured as follows. Section \ref{sec:related_works} describes the related literature. Section \ref{sec:datasets} describes the dataset used. Section \ref{sec:model} describes selected state-of-the-art deep learning models for forecasting. Section \ref{sec:methodology} describes the methodology of the comparison analysis. Section \ref{sec:results} describes the results. Section \ref{sec:findings} discusses the research questions and the lessons learned. Finally, section \ref{sec:conclusion} summarizes the paper.

\section{Related works} \label{sec:related_works}

Network traffic forecasting is a critical area of research due to its importance in network optimization, resource allocation, and anomaly detection. Various approaches have been developed over the years, leveraging both statistical and machine learning methods.

Currently, statistically-based methods are commonly applied in network traffic forecasting with many published approaches. Such statistically-based methods include Auto-Regressive Integrated Moving Average (ARIMA), Seasonal ARIMA (SARIMA), Exponential Smoothing (ETS), or Hidden Markov Models \cite{moayedi2008arima,moussas2005network,tikunov2007traffic,chen2016predicting}. However, there are several limitations of statistical-based forecasting \cite{ferreira2023forecasting,khashei2011methodology,clements1993limitations}. Most of them rely on the assumption of linearity, which restricts their ability to model complex, non-linear relationships often present in real-world systems. Methods like ARIMA and regression models require stationarity in data, meaning trends and variance must remain constant over time, necessitating preprocessing that can complicate analysis. Moreover, these models are sensitive to parameter estimation, and their predictive accuracy can degrade in highly dynamic environments where relationships between variables fluctuate. Statistical approaches also struggle to handle multivariate interactions effectively, leading to oversimplified representations of intricate patterns. These limitations collectively underscore the challenges of applying statistical models in scenarios with complex, non-linear, or rapidly changing dynamics.

Therefore, deep learning has emerged in recent years as a transformative approach to network traffic forecasting. Primarily due to its ability to model non-linear relationships, capture complex dependencies within data, handle non-stationary time series effectively, and provide superior scalability \cite{ferreira2023forecasting,joshi2015review}.  

Deep learning for forecasting became increasingly valuable for forecasting, particularly with the introduction of Long Short-Term Memory (LSTM) and Gated Recurrent Units (GRU),which mitigate the exploding and vanishing gradient problems that affected earlier Recurrent Neural Networks (RNN)~\cite{lim2021time}. These deep learning models have been successfully applied in the network traffic forecasting domain~\cite{dalgkitsis2018traffic,ramakrishnan2018network}. Nevertheless, the rapid expansion of deep learning forecasting variants in recent years underscores the need for a comprehensive evaluation.

The first major enhancement of LSTM and GRU models was their integration with Fully Convolutional Networks (FCNs)~\cite{karim2017lstm,elsayed2018deep}. These hybrid models gained popularity due to their ability to outperform classical approaches while requiring minimal data preprocessing and easily scaling to large time series datasets~\cite{karim2019insights}.

Moreover, more sophisticated deep learning models originally developed for image classification have also been applied in time series forecasting area. Notable examples include Residual Networks (ResNet) \cite{he2016deep}. Building on these architectures, additional models have been developed, such as InceptionTime~\cite{inceptiontime}, which integrates residual blocks with Global Average Pooling layers.

Finally, one of the most recent forecasting approaches uses Transformers, which have become a prominent tool in time series forecasting due to their capability to model long-range dependencies, critical for network traffic prediction~\cite{wen2022transformers}. However, traditional Transformers face challenges such as inefficiency and a lack of interpretability in capturing temporal dynamics. To address these limitations, several advancements have been proposed. For instance, ETSformer~\cite{woo2022etsformer} introduces exponential smoothing and frequency attention mechanisms to replace conventional self-attention, enabling the model to capture both seasonal and trend components in a decomposable and interpretable manner. Similarly, models like Informer and Autoformer leverage sparsity and frequency-based approaches to enhance efficiency and adaptability in long-term sequence forecasting. Despite their success, recent studies question the necessity of complex Transformer-based architectures, with findings that simple linear models can outperform them in several scenarios~\cite{zeng2023transformers}. Nevertheless, Transformers remain at the forefront of innovations in network traffic forecasting due to their ability to incorporate dynamic time-dependent features and extract meaningful patterns from large-scale, non-stationary data.

In summary, related works establish a solid foundation for future research and practical applications by tackling the challenges of non-linearity and uncertainty in traffic patterns. However, a major limitation persists in the datasets used for evaluation.

Many of the published works utilize different datasets which makes approaches hardly comparable. For example, many datasets used in the network traffic forecasting domain were not created with the intention of forecasting evaluation, making them unsuitable for such purposes. An example of this is WIDE project data in the MAWI archive \cite{mawilab}, which contains only 15 minutes per day. Moreover, some related works are built on datasets that are not publicly available \cite{do2020prediction,zhou2012predictability}, making it impossible to independently verify the results. Others rely on synthetic data \cite{8840407,li2022network}, which, while useful for controlled experiments, fail to provide insights into real-world network environments. Furthermore, even widely used datasets, such as KDD 99 and NSL-KDD, may lack the comprehensive representation of modern, dynamic traffic patterns required for robust evaluation.

The recent release of a large, standardized, and publicly available real-world dataset CESNET-TimeSeries24 from an ISP network provides a valuable resource for comparing existing forecasting models. This comprehensive dataset enables a more consistent evaluation of deep learning approaches in realistic network scenarios. By leveraging this dataset, our approach aims to perform a comparative analysis of state-of-the-art deep learning models, filling a crucial gap in the domain and fostering more reliable and meaningful performance assessments.

\section{Dataset} \label{sec:datasets}

To evaluate state-of-the-art deep learning models for real-world network traffic forecasting, we selected a dataset from an actual operational network. The CESNET-TimeSeries24 dataset \cite{koumar2024cesnettimeseries24timeseriesdataset} was chosen for this purpose, as it captures traffic data from the ISP network CESNET3\footnote{The Czech Educational and Science Network}.

The dataset is uniquely comprehensive, offering multivariate time series created through traffic aggregation at three distinct intervals: 10\,minutes, 1\,hour, and 1\,day. Each interval-specific time series contains multiple metrics useful for network traffic forecasting and anomaly detection, as listed in Table \ref{tab:attributes}. Given that the original capture interval was 10\,minutes, metrics that represent unique values were adapted for the 1\,hour and 1\,day series incorporating sum, average and standard deviation statistics to enhance temporal granularity.

\begin{table}[h]
\centering
\caption{Time series metrics of time series from the dataset.}\label{tab:attributes}
\begin{tabular}{p{4cm}|p{8cm}}
\toprule
\textbf{Time series metric name} & \textbf{Description}         \\ 
\toprule
n\_flows                                    & The number of IP flows                                   \\ 
\rowcolor{LigthGray} n\_packets                                  & The number of packets                                 \\ 
n\_bytes                                    & The number of bytes                                   \\
\rowcolor{LigthGray} n\_dest\_ip                           & The number of unique destination IPs                           \\ 
n\_dest\_ports                       & The number of unique destination ports                \\ 
\rowcolor{LigthGray} n\_dest\_asn                           & The number of unique destination ASNs  \\ 
tcp\_udp\_ratio\_packets                    & The TCP/UDP packet ratio                           \\ 
\rowcolor{LigthGray} tcp\_udp\_ratio\_bytes                      & The TCP/UDP bytes ratio                            \\ 
dir\_ratio\_packets                         & The directional ratio of packets                      \\ 
\rowcolor{LigthGray} dir\_ratio\_bytes                           & The directional ratio of bytes                        \\ 
avg\_duration                               & The average duration                                  \\ 
\rowcolor{LigthGray} avg\_ttl                                    & The average TTL                                       \\ 
\bottomrule
\end{tabular}
\end{table}

A standout feature of the CESNET-TimeSeries24 dataset is its extensive coverage and granularity, encompassing not only general network traffic, but also highly detailed time series across 283 institutions, 548 institutional subnets and over 270,000 individual IP addresses. This extensive range provides a robust basis for comparative analysis of deep learning models, enabling researchers to benchmark forecasting performance across multiple hierarchical levels within the network. This dataset enables a detailed examination of the model's scalability, adaptability, and performance in forecasting diverse traffic patterns within a real-world ISP environment.

Spanning a substantial 40-week period, from October 2023 to July 2024, the dataset captures both long-term trends and fine-grained fluctuations, offering an invaluable resource for rigorous deep-learning model assessment.

\section{Deep Learning models for forecasting} \label{sec:model}

This section provides a brief description of model types or particular state-of-the-art deep learning model architectures chosen to be compared in this study. The types were chosen to represent versatility in time series prediction handling tasks and to vary in computational demands.

Moreover, we provide information about their general advantages and shortcomings that can be found in related works. These models were selected due to their distinct architectural approaches, allowing for a comparative analysis of their strengths and limitations in different predictive contexts. Given the diverse requirements of real-time network operations, the models range from resource-intensive deep learning architectures to more streamlined alternatives, each offering unique benefits and trade-offs. Understanding these differences is essential to assess the deployability of models in real-world network environments, where both computational efficiency and prediction accuracy are critical factors.

\begin{description}
    \item[Mean] model used as a baseline for evaluating deep learning models, computes the average value of datapoints from the training window and applies this mean as the prediction for each datapoints in the prediction window. While its simplicity is advantageous for benchmarking, it can capture only trends and fails to capture temporal patterns and seasonality, making it unsuitable for datasets with significant time dependencies. Nevertheless, its role as a baseline is crucial, as it sets a minimal performance threshold that deep learning models should surpass, ensuring their predictive capacity is not inferior to a simplistic, non-adaptive approach.

    \item[Long Short-Term Memory (LSTM)] networks are well-regarded for their capacity to capture long-term dependencies in sequence data, as they mitigate the vanishing gradient problem that hinders traditional RNNs. The architecture utilizes memory cells, which allow the model to retain information over extended time steps, making it highly effective for time-series prediction tasks. However, LSTMs are computationally intensive and require substantial memory resources, which can be a limitation in environments with constrained computational power. Additionally, they tend to have a large parameter count, which can lead to overfitting, especially when working with smaller datasets.

    \item[Gated Recurrent Unit (GRU)] networks are a simplified alternative to LSTM networks, designed to efficiently capture dependencies in sequential data while addressing the vanishing gradient problem. GRUs streamline the architecture by combining the forget and input gates into a single update gate, which reduces the parameter count and computational demands. This makes GRUs faster to train and more memory-efficient, especially on smaller datasets or resource-constrained environments. However, their simplified structure may limit their ability to model complex long-term dependencies as effectively as LSTMs, which can be a drawback for tasks requiring extended temporal context.

    \item[LSTM-FCN] model \cite{karim2017lstm} combines the strengths of LSTM networks with Fully Convolutional Networks (FCNs), enabling the model to capture both temporal and spatial features in time-series data. This hybrid architecture improves feature extraction and helps generalize better to complex data patterns. Nevertheless, the increased complexity of combining LSTM with FCN layers results in greater computational demands and can make training slower, especially on larger datasets or limited hardware.

    \item[GRU-FCN] model \cite{elsayed2018deep} integrates GRUs with Fully Convolutional Networks, blending the benefits of a simpler recurrent architecture with powerful feature extraction capabilities. GRUs offer a streamlined alternative to LSTMs, potentially reducing training time due to their more efficient structure. Nevertheless, GRU-FCN may not capture long-term dependencies with the same depth as LSTM models, which can be a drawback when handling complex, long-sequence data.

    \item[ResNet] models \cite{he2016deep} are known for their residual connections, which help mitigate the vanishing gradient problem in deep neural networks, making them well-suited for extracting hierarchical features in time-series data. The residual connections enable training deeper architectures without the degradation issues typical in very deep networks. Nonetheless, ResNets can be resource-intensive and may require a significant amount of data to reach their full potential, as they are prone to overfitting on smaller datasets.
    
    \item[InceptionTime] model \cite{inceptiontime} is a deep convolutional model inspired by the Inception architecture, specifically adapted for time-series classification. Its multi-scale feature extraction capabilities allow it to capture both local and global patterns in time-series data, contributing to its high performance across various benchmark tasks. However, the model’s depth and complexity make it computationally expensive and may require extensive tuning to achieve optimal results, especially in cases with less structured or lower-dimensional data.

    \item[Random Connectivity LSTM (RCLSTM)] model \cite{rslstm} is an optimized variant of LSTM designed to reduce computational complexity while maintaining comparable predictive performance. RCLSTM introduces stochastic connectivity to conventional LSTM neurons, creating a sparse neural structure that significantly decreases the number of trainable parameters. This modification improves computational efficiency, making the model particularly suitable for latency-stringent and power-constrained environments. Despite the reduction in connections, RCLSTM preserves long-range dependencies effectively, making it highly applicable in tasks such as traffic forecasting and user mobility prediction.
\end{description}

The models and architectures discussed in this section are commonly used in the time series prediction domain, each providing promising robustness and modelling potential for this task, while varying in computational complexity. The variability in the chosen models should demonstrate the potential for their application in real-world network operations, where one has to balance between accuracy and computational efficiency.

\section{Methodology} \label{sec:methodology}

Our methodology follows the recommended usage of the CESNET-TimeSeries24 dataset~\cite{koumar2024cesnettimeseries24timeseriesdataset}. We use multiple dataset parts provided in the dataset---institutions, institution subnets, and an IP address sample. We trained models on time series aggregated on one-hour time intervals, and each time series metric was modelled separately. Thus, we provide a comparison of only univariate approaches. We evaluate models on all time series metrics available in the dataset.

\begin{figure}[h]
    \centering
    \includegraphics[width=0.55\linewidth]{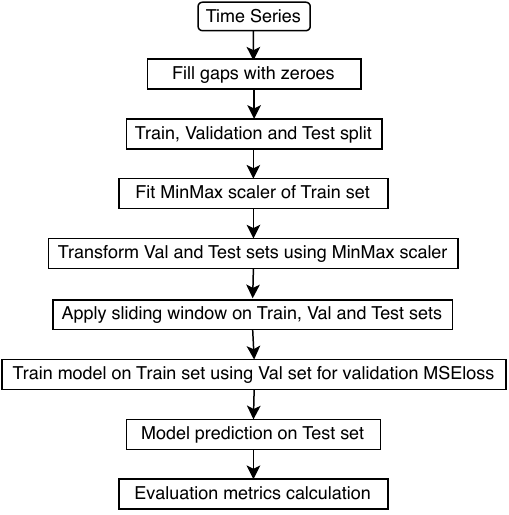}
    \caption{Used methodology applied on each time series metric}
    \label{fig:methodology}
\end{figure}

Pre-processing involves several steps for each time series, as shown in Figure~\ref{fig:methodology}, including data splitting, standardizing the data, and addressing missing values.

\subsection{Missing values handling}

Real-world network traffic is usually not constantly generated, especially in the case of monitoring a single device. Thus, it is inevitable that the collected time series contains missing values~\cite{koumar2023unevenly}, which was identified as one of the challenges of the dataset~\cite{koumar2024cesnettimeseries24timeseriesdataset}.

In this comparative study, we fill missing values with zeros, which ideally represents zero transmitted traffic (IP flows, packets, bytes, etc.). Therefore, one of the objectives of this study is to verify how DL models handle time series with filled zero values that, in extreme cases, can represent a majority of values in particular time series.

\subsection{Time series splitting, normalization, and sliding window}

Each time series is split into train, validation, and test sets for the correct model training and evaluation. The duration of the training set represents the first 35\% of the time series (14 weeks). The validation set holds the next 5\% (2 weeks), and the remainder (24 weeks) is used as the test set. Notably, none of the models are retrained during evaluation on the Test set, providing a clear advantage over iterative approaches such as SARIMA. 

At first, we fill in missing time series values with zeros. Then, we train the MinMax scaler on the train set and transform the train, validation, and test parts of the data using the fitted scaler.

We used the sliding-window approach, which is a widely adopted method for splitting data into multiple training samples from a single continuous time series stream. In our methodology, each neural network has an input of size equal to a number of consecutive time series values in a training window (also often referred to as a look-back window) of fixed length from which the model learns temporal dependencies. Based on this input, the model generates a specified number of consecutive future values. The number of predicted future values depends on a predefined prediction window size (also often referred to as a forecast horizon). The specific combinations of the training window size and the prediction window size evaluated in this study are summarized in Table~\ref{tab:windows}. We have chosen training and prediction window sizes commonly used in related works. By systematically adjusting the training and prediction window sizes---such as using daily, weekly, and monthly lookback spans along with hourly, daily, and weekly forecast horizons---we explore how the models' performance varies depending on the amount of historical data considered and the length of the forecast.

\begin{table}[h]
    \centering
    \caption{Used training and prediction windows}\label{tab:windows}
    \begin{tabular}{c|c}
        \toprule
        \textbf{Training window size} & \textbf{Prediction window size} \\
        \toprule
         24  & 1 \\ 
         168 & 1 \\ 
         168 & 24 \\ 
         744 & 1 \\ 
         744 & 168 \\
        \bottomrule
    \end{tabular}
\end{table}

\begin{figure}[h]
    \centering
    \includegraphics[width=\linewidth]{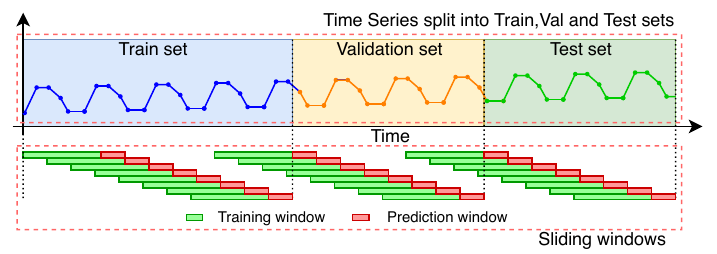}
    \caption{Illustration of time series partitioning into training, validation, and test sets, along with the sliding window approach for generating overlapping training and prediction windows.}
    \label{fig:timeseries-split}
\end{figure}

Once a training–prediction pair is formed, the window is shifted forward by a prediction window. This process is repeated across the entire dataset, creating overlapping sequences that capture both short-term and long-term variations in the data. As a result, the model is trained on diverse input-output examples rather than relying on a single continuous segment, thereby improving its ability to generalize.
In this study, the sliding window approach is applied consistently to the training, validation, and test sets of the time series, as illustrated in Figure~\ref{fig:timeseries-split}. 

\subsection{Models optimization}

During training, the Adam optimizer was employed, with MSELoss serving as the loss function to measure discrepancies between predicted and actual values. A hyper-parameter search was conducted to explore various architectural and optimization configurations, including the number of RNN layers, hidden layer size, batch size, learning rate, and number of epochs. The tested number of RNN layers and hidden size were chosen for each model based on related works that use the model for network traffic prediction. Most of the models were trained with 1 to 5 RNN layers and hidden layer sizes from 20 to 200. We perform a searching for optimal architecture on this scale of the number and sizes of layers. The final architecture and hyper-parameter settings, which represent a balance between predictive accuracy and computational efficiency, are presented in Table~\ref{tab:hyperparam}.

\begin{table}[h]
    \centering
    \caption{Summary of Final Hyperparameters}\label{tab:hyperparam}
    \begin{tabular}{l|l|l|l|l|l}
        \toprule
        \multicolumn{1}{l|}{\textbf{Model}}  & \textbf{RNN layers} & \textbf{Hidden size} & \textbf{Batch} & \textbf{Learning rate} & \textbf{Epochs} \\
        \toprule
        GRU & 1 & 100 & 16 & 0.01 & 100 \\ 
        LSTM & 1 & 100 & 16 & 0.01 & 100 \\ 
        GRU-FCN & 4 & 20 & 16 & 0.01 & 20 \\ 
        LSTM-FCN & 1 & 100 & 16 & 0.01 & 100 \\ 
        InceptionTime & - & - & 128 & 0.001 & 20 \\
        ResNet & - & - & 128 & 0.01 & 20 \\
        RCLSTM & 1 & 300 & 32 & 0.01 & 100 \\ 
        \bottomrule
    \end{tabular}
\end{table}

These hyper-parameters were selected based on mean results on a random sample of 50 time series from each dataset part. Most of the models ended with one RNN layer, which may be surprising at first glance. However, it is in line with the findings of Zeng et al. \cite{zeng2023transformers}. In their work, they prove that a simple model can outperform even transformers. For the LSTM-FCN and GRU-FCN models, we also tuned the sizes of CNN layers. Both model architectures contain three CNN layers with resulting best-performing sizes 128, 256, and 128 for LSTM-FCN, and 64-128-64 for GRU-FCN.

For the RCLSTM model, the architecture and hyperparameters were taken from the original publication as their methodology also performs hourly forecasting. Likewise, InceptionTime and ResNet were deployed with their standard, predefined architectures. By consolidating and refining the results of multiple experimental trials, the chosen hyperparameters perform best from evaluated across a range of time series in used datasets parts, as presented in the next section.

\subsection{Evaluation measures}

The evaluation follows a common procedure where we forecast the next values based on previous ones using the trained models. To evaluate the precision of the forecasting, we use two fully unsupervised measures, Root Mean Square Error (RMSE) and $R^2$-score (also called the coefficient of determination). In the following equations, $y$ represents the actual normalized value, while $\hat{y}$ represents the predicted value. The RMSE is defined as:

\begin{equation}
 \text{RMSE}(y, \hat{y}) = \sqrt{\frac{1}{N} \sum_{i=0}^{N - 1} (y_i - \hat{y}_i)^2}.
\end{equation}

The $\text{RMSE}$ ranges from \(0\) to \(\infty\), with lower values indicating more accurate predictions. The measure highlights models that do not have large errors in prediction. However, its sensitivity to large errors can be insufficient, sometimes obscuring the distinction between consistent underperformance and occasional extreme errors.

The $R^2$-score measure (also called the coefficient of determination) is a statistical measure that shows the proportion of the dependent variable's variance explained by the independent variable, calculated using the following equation:

\begin{equation}
    R^2(y, \hat{y}) = 1 - \frac{\sum_{i=1}^n (y_i - \hat{y}_i)^2}{\sum_{i=1}^n (y_i - \mu_y)^2  + \epsilon}.   \label{r2score}
\end{equation}

The \( R^2 \)-score ranges from \(-\infty\) to \(1\), with higher values indicating more accurate predictions. The $R^2$-score measure estimates goodness-of-fit, showing how well the model captures overall data trends.

Combining RMSE and $R^2$-score measures offers a comprehensive evaluation of forecasting models by capturing different aspects of performance. RMSE emphasizes large individual errors and highlights variance, while $R^2$-score reflects how well the model explains the variance in the dependent variable.

To balance these contributions, we introduce a combined measure called the \textit{Harmonic-Score}, which uses the harmonic mean to penalize extreme values and ensure both metrics influence the final score similarly. Since RMSE and $R^2$-score operate on different scales, we rescale them by clipping RMSE at a maximum of 11 and $R^2$ at a minimum of -10. These thresholds were chosen based on the distribution of model results, capturing the central $95\%$ of the data to limit the influence of outliers while preserving typical values. This adjustment allows for a meaningful combination using the harmonic mean without distorting the overall evaluation. The \textit{Harmonic-Score} measure is defined as:

\begin{equation}
%\[
\text{\textit{Harmonic-Score}} = 2 \frac{\text{RMSE} \left| \text{R}^2 - 1\right|}{\text{RMSE} + \left|\text{R}^2 - 1\right|}.
%\]
\end{equation}

This equation adjusts the \(R^2\)-score by transforming it to align with RMSE, ensuring both measures are compatible. The harmonic mean emphasizes instances where both measures indicate precision, offering a clear, balanced evaluation.

\section{Experimental forecasting results} \label{sec:results}

We evaluate the selected DL models across each time series metric and dataset, leading to a large volume of results. However, due to space limitations, not all results can be presented in the paper. Therefore, we focus on presenting the results of the metric \textit{Number of bytes} (transmitter data), as it is commonly used in the network traffic forecasting domain. The results for the remaining features are presented briefly, focusing on key insights. All results and analyses are available in the supplementary materials on the GitHub repository\footnote{\url{https://github.com/koumajos/isp-forecasting-benchmark}\label{footnote:github}}.

\begin{table*}
\centering
\caption{Resulting mean and standard deviation of RMSE in the format ``\(<\)mean\(>\) (\(<\)standard deviation\(>\))".}\label{tab:RMSE}
\begin{tabular}{r|r|r|C{1.5cm}|C{1.5cm}|C{1.5cm}|C{1.5cm}|C{1.5cm}|C{1.5cm}|C{1.5cm}|C{1.5cm}}
\toprule
 & \rotatebox{90}{\textbf{Training window}} & \rotatebox{90}{\textbf{Prediction \ window}}  & \rotatebox{90}{\textbf{Mean}}  & \rotatebox{90}{\textbf{GRU}}  & \rotatebox{90}{\textbf{LSTM}}  & \rotatebox{90}{\textbf{GRU-FCN}}  & \rotatebox{90}{\textbf{LSTM-FCN}}  & \rotatebox{90}{\textbf{InceptionTime}}  & \rotatebox{90}{\textbf{ResNet}}  & \rotatebox{90}{\textbf{RCLSTM}}  \\
\toprule

  & 24  & 1 & \cellcolor{LightRed}0.146 (0.75)& 0.104 (0.53)& 0.105 (0.54)& \cellcolor{LighterGreen}0.102 (0.55) & \cellcolor{LightGreen}\textbf{0.102 (0.54)} & 0.112 (0.55)& 0.106 (0.55)& \cellcolor{LighterRed}0.140 (0.8)\\
  
  & 168  & 1 & \cellcolor{LightRed}0.149 (0.75)& 0.105 (0.53)& 0.106 (0.54)& \cellcolor{LighterGreen}0.104 (0.55) & \cellcolor{LightGreen}\textbf{0.103 (0.54)} & 0.124 (0.55)& 0.127 (0.55)& \cellcolor{LighterRed}0.140 (0.8)\\
  
  & 168  & 24 & 0.150 (0.75)& 0.123 (0.55)& 0.124 (0.55) & \cellcolor{LightGreen}\textbf{0.115 (0.55)} & \cellcolor{LighterGreen}0.117 (0.55)& \cellcolor{LightRed}0.483 (0.51)& 0.131 (0.55)& \cellcolor{LighterRed}0.163 (0.81)\\
  
  & 744  & 1 & \cellcolor{LightRed}0.151 (0.75)& \cellcolor{LightGreen}\textbf{0.106 (0.53)} & \cellcolor{LightGreen}\textbf{0.106 (0.53)} & 0.111 (0.55)& \cellcolor{LighterGreen}0.110 (0.53)& 0.136 (0.55)& \cellcolor{LighterRed}0.146 (0.56)& 0.112 (0.58)\\
  
\parbox[t]{2mm}{\multirow{-5}{*}{\rotatebox[origin=c]{90}{\textbf{Inst.}}}}  & 744  & 168  & \cellcolor{LightGreen}\textbf{0.152 (0.76)} & 0.165 (0.82)& 0.165 (0.82)& 0.165 (0.82)& \cellcolor{LighterGreen}0.164 (0.82)& \cellcolor{LightRed}0.519 (0.78)& 0.173 (0.82)& \cellcolor{LighterRed}0.193 (0.83)\\
    
    \bottomrule

   %\multicolumn{3}{r|}{\textbf{Overall}}  & \cellcolor{LighterRed}0.153 (0.8) &  0.121 (0.61) &  0.122 (0.62) &  \cellcolor{LightGreen}\textbf{0.120 (0.62)} &  \cellcolor{LighterGreen}0.120 (0.62) &  \cellcolor{LightRed}0.276 (0.63) &  0.137 (0.63) &  0.152 (0.8) \\
   \multicolumn{3}{r|}{\textbf{Mean}}  & \cellcolor{LighterRed}0.1496 & 0.1206 & 0.1212 & \cellcolor{LighterGreen}0.1194 & \cellcolor{LightGreen}\textbf{0.1192} & \cellcolor{LightRed}0.2748 & 0.1366 & \cellcolor{LighterRed}0.1496 \\
\bottomrule

  & 24  & 1 & \cellcolor{LightRed}0.383 (1.63)& 0.218 (1.14)& 0.219 (1.15) & \cellcolor{LightGreen}\textbf{0.217 (1.15)} & \cellcolor{LighterGreen}0.217 (1.15)& 0.230 (1.15)& 0.220 (1.15)& \cellcolor{LighterRed}0.323 (1.5)\\
  
  & 168  & 1 & \cellcolor{LightRed}0.392 (1.66) & \cellcolor{LightGreen}\textbf{0.219 (1.15)} & 0.220 (1.15)& \cellcolor{LighterGreen}0.219 (1.15)& 0.219 (1.15) & 0.238 (1.15)& 0.235 (1.15)& \cellcolor{LighterRed}0.277 (1.36)\\
  
  & 168  & 24 & \cellcolor{LighterRed}0.394 (1.66)& 0.237 (1.15)& 0.237 (1.15) & \cellcolor{LightGreen}\textbf{0.229 (1.15)} & \cellcolor{LighterGreen}0.231 (1.15)& \cellcolor{LightRed}0.59 (1.1)& 0.244 (1.15)& 0.334 (1.48)\\
  
  & 744  & 1 & \cellcolor{LightRed}0.400 (1.68) & \cellcolor{LightGreen}\textbf{0.220 (1.14)} & \cellcolor{LighterGreen}0.220 (1.15)& 0.228 (1.15)& 0.228 (1.15)& 0.253 (1.15)& \cellcolor{LighterRed}0.255 (1.15)& 0.225 (1.16)\\
  
\parbox[t]{2mm}{\multirow{-5}{*}{\rotatebox[origin=c]{90}{\textbf{Inst. subnets}}}}   & 744  & 168 & \cellcolor{LighterRed}0.402 (1.69)& \cellcolor{LighterGreen}0.265 (1.23) & \cellcolor{LightGreen}\textbf{0.265 (1.23)} & 0.269 (1.23)& 0.265 (1.23)& \cellcolor{LightRed}0.613 (1.18)& 0.275 (1.23)& 0.326 (1.37)\\

    \bottomrule

   %\multicolumn{3}{r|}{\textbf{Overall}}  & \cellcolor{LightRed}0.419 (1.81) &  \cellcolor{LightGreen}\textbf{0.243 (1.26)} &  0.244 (1.26) &  \cellcolor{LighterGreen}0.244 (1.26) &  0.244 (1.26) &  \cellcolor{LighterRed}0.396 (1.25) &  0.257 (1.26) &  0.314 (1.5) \\
   \multicolumn{3}{r|}{\textbf{Mean}}  & \cellcolor{LightRed}0.3942 & \cellcolor{LightGreen}\textbf{0.2318} & 0.2322 & 0.2324 & \cellcolor{LighterGreen}0.2320 & \cellcolor{LighterRed}0.3848 & 0.2458 & 0.297 \\
   
\bottomrule

  & 24  & 1 & \cellcolor{LightRed}1.01 (2.86) & \cellcolor{LightGreen}\textbf{0.149 (0.82)} & \cellcolor{LighterGreen}0.150 (0.82)& 0.150 (0.82)& 0.151 (0.82)& 0.165 (0.82)& 0.152 (0.82)& \cellcolor{LighterRed}0.221 (1.08)\\
  
  & 168  & 1 & \cellcolor{LightRed}1.011 (2.86) & \cellcolor{LightGreen}\textbf{0.150 (0.82)} & \cellcolor{LighterGreen}0.151 (0.82)& 0.152 (0.82)& 0.152 (0.82)& 0.168 (0.82)& 0.158 (0.82)& \cellcolor{LighterRed}0.226 (1.12)\\
  
  & 168  & 24 & \cellcolor{LightRed}1.011 (2.86)& 0.154 (0.82)& \cellcolor{LighterGreen}0.154 (0.82) & \cellcolor{LightGreen}\textbf{0.154 (0.82)} & 0.158 (0.82)& \cellcolor{LighterRed}0.573 (0.77)& 0.159 (0.82)& 0.247 (1.17)\\
  
  & 744  & 1 & \cellcolor{LightRed}1.013 (2.86) & \cellcolor{LightGreen}\textbf{0.150 (0.82)} & \cellcolor{LighterGreen}0.151 (0.82)& 0.153 (0.82)& 0.154 (0.82)& \cellcolor{LighterRed}0.183 (0.82)& 0.163 (0.82)& 0.166 (0.89)\\
  
\parbox[t]{2mm}{\multirow{-5}{*}{\rotatebox[origin=c]{90}{\textbf{IP addr.}}}}   & 744  & 168 & \cellcolor{LightRed}1.014 (2.86)& \cellcolor{LighterGreen}0.179 (0.93) & \cellcolor{LightGreen}\textbf{0.179 (0.93)} & 0.214 (0.92)& 0.189 (0.93)& \cellcolor{LighterRed}0.593 (0.88)& 0.182 (0.93)& 0.234 (1.08)\\

    \bottomrule

   %\multicolumn{3}{r|}{\textbf{Overall}}  & \cellcolor{LightRed}1.102 (3.14) &  \cellcolor{LightGreen}\textbf{0.161 (0.9)} &  \cellcolor{LighterGreen}0.162 (0.9) &  0.169 (0.9) &  0.166 (0.9) &  \cellcolor{LighterRed}0.341 (0.9) &  0.168 (0.9) &  0.228 (1.16)  \\
   \multicolumn{3}{r|}{\textbf{Mean}}  & \cellcolor{LightRed}1.0098 & \cellcolor{LightGreen}\textbf{0.1564} & \cellcolor{LighterGreen}0.1570 & 0.1646 & 0.1608 & \cellcolor{LighterRed}0.3364 & 0.1628 & 0.2188 \\

\bottomrule

   \multicolumn{3}{r|}{\textbf{Overall mean}}  & \cellcolor{LightRed}0.5179 & \cellcolor{LightGreen}\textbf{0.1696} & \cellcolor{LighterGreen}0.1701 & 0.1721 & 0.1707 & \cellcolor{LighterRed}0.332 & 0.1817 & 0.2218 \\
 % \multicolumn{3}{r|}{\textbf{Overall}}  & \cellcolor{LightRed}0.75 (2.57) & \cellcolor{LightGreen}\textbf{0.18 (0.99)} & \cellcolor{LighterGreen}0.181 (0.99) & 0.185 (0.99) & 0.182 (0.99) & \cellcolor{LighterRed}0.348 (0.99) & 0.19 (0.99) & 0.242 (1.23) \\

% \bottomrule

\end{tabular}
\end{table*}

\begin{table*}
\centering
\caption{Resulting mean and standard deviation of $R^2\text{-score}$ in the format ``\(<\)mean\(>\) (\(<\)standard deviation\(>\))".}\label{tab:R2-SCORE}
\begin{tabular}{r|r|r|C{1.5cm}|C{1.5cm}|C{1.5cm}|C{1.5cm}|C{1.5cm}|C{1.5cm}|C{1.5cm}|C{1.5cm}}
& \rotatebox{90}{\textbf{Training window}} & \rotatebox{90}{\textbf{Prediction \ window}}  & \rotatebox{90}{\textbf{Mean}}  & \rotatebox{90}{\textbf{GRU}}  & \rotatebox{90}{\textbf{LSTM}}  & \rotatebox{90}{\textbf{GRU-FCN}}  & \rotatebox{90}{\textbf{LSTM-FCN}}  & \rotatebox{90}{\textbf{InceptionTime}}  & \rotatebox{90}{\textbf{ResNet}}  & \rotatebox{90}{\textbf{RCLSTM}}  \\
\toprule

  & 24  & 1 & 0.09 (0.1)& 0.08 (0.8)& 0.09 (0.8)& 0.15 (1.0)& \cellcolor{LighterGreen}0.19 (0.7)& \cellcolor{LightRed}-0.77 (2.5)& \cellcolor{LighterRed}0.06 (1.1) & \cellcolor{LightGreen}\textbf{0.2 (0.9)} \\
  
  & 168  & 1 & 0.04 (0.1)& 0.03 (0.9)& 0.1 (0.7)& \cellcolor{LighterGreen}0.17 (0.7)& 0.14 (0.9)& \cellcolor{LightRed}-1.08 (2.5)& \cellcolor{LighterRed}-0.63 (2.2) & \cellcolor{LightGreen}\textbf{0.2 (0.9)} \\
  
  & 168  & 24  & \cellcolor{LightGreen}\textbf{0.03 (0.1)} & -0.45 (1.2)& -0.45 (1.3)& \cellcolor{LighterGreen}-0.18 (1.1)& -0.29 (1.3)& \cellcolor{LightRed}-9.22 (2.2)& \cellcolor{LighterRed}-0.73 (1.7)& -0.37 (1.5)\\
  
  & 744  & 1 & 0.01 (0.1)& 0.06 (0.7)& \cellcolor{LighterGreen}0.07 (0.8)& -0.05 (1.0)& -0.09 (1.2)& \cellcolor{LightRed}-1.59 (2.8)& \cellcolor{LighterRed}-0.92 (2.4) & \cellcolor{LightGreen}\textbf{0.23 (0.7)} \\
  
\parbox[t]{2mm}{\multirow{-5}{*}{\rotatebox[origin=c]{90}{\textbf{Inst.}}}}    & 744  & 168  & \cellcolor{LightGreen}\textbf{-0.01 (0.1)} & -0.53 (1.3)& \cellcolor{LighterGreen}-0.49 (1.2)& -1.08 (2.8)& -0.54 (1.4)& \cellcolor{LightRed}-9.18 (2.3)& -0.74 (1.5)& \cellcolor{LighterRed}-1.8 (2.4)\\

\bottomrule

 %\multicolumn{3}{r|}{\textbf{Overall}}  &  \cellcolor{LighterGreen}0.03 (0.1) &  -0.04 (0.4) &  -0.03 (0.4) &  \cellcolor{LightGreen}\textbf{0.05 (0.4)} &  0.02 (0.4) &  \cellcolor{LightRed}-0.57 (0.5) &  \cellcolor{LighterRed}-0.19 (0.4) &  0.0 (0.7) \\
   \multicolumn{3}{r|}{\textbf{Mean}}  & \cellcolor{LightGreen}\textbf{0.032} & -0.162 & -0.136 & -0.198 & \cellcolor{LighterGreen}-0.118 & \cellcolor{LightRed}-4.368 & \cellcolor{LighterRed}-0.592 & -0.308 \\

\bottomrule

  & 24  & 1  & \cellcolor{LightGreen}\textbf{0.12 (0.2)} & -0.1 (1.3)& -0.05 (1.1)& 0.06 (0.8)& \cellcolor{LighterGreen}0.08 (0.8)& \cellcolor{LightRed}-1.29 (3.1)& \cellcolor{LighterRed}-0.15 (1.4)& 0.08 (1.1)\\
  
  & 168  & 1 & \cellcolor{LighterGreen}0.06 (0.1)& -0.1 (1.2)& -0.07 (1.2)& 0.05 (0.9)& -0.01 (1.2)& \cellcolor{LightRed}-1.57 (3.0)& \cellcolor{LighterRed}-0.68 (2.1) & \cellcolor{LightGreen}\textbf{0.09 (1.0)} \\
  
  & 168  & 24  & \cellcolor{LightGreen}\textbf{0.05 (0.1)} & -0.55 (1.4)& -0.53 (1.3)& \cellcolor{LighterGreen}-0.27 (1.1)& -0.55 (1.8)& \cellcolor{LightRed}-8.9 (2.7)& \cellcolor{LighterRed}-0.77 (1.6)& -0.64 (1.9)\\
  
  & 744  & 1 & \cellcolor{LighterGreen}0.02 (0.1)& -0.1 (1.2)& -0.09 (1.1)& -0.19 (1.1)& -0.24 (1.3)& \cellcolor{LightRed}-2.19 (3.4)& \cellcolor{LighterRed}-1.05 (2.5) & \cellcolor{LightGreen}\textbf{0.1 (1.0)} \\
  
\parbox[t]{2mm}{\multirow{-5}{*}{\rotatebox[origin=c]{90}{\textbf{Inst. subnets}}}}   & 744  & 168  & \cellcolor{LightGreen}\textbf{0.0 (0.1)} & -0.61 (1.4)& \cellcolor{LighterGreen}-0.55 (1.3)& -1.67 (3.3)& -0.94 (2.2)& \cellcolor{LightRed}-8.85 (2.8)& -0.91 (1.8)& \cellcolor{LighterRed}-2.26 (2.9)\\

\bottomrule

 %\multicolumn{3}{r|}{\textbf{Overall}}  &  \cellcolor{LightGreen}\textbf{0.05 (0.1)} &  -0.1 (0.4) &  -0.09 (0.4) &  \cellcolor{LighterGreen}-0.05 (0.4) &  -0.06 (0.4) &  \cellcolor{LightRed}-0.61 (0.5) &  \cellcolor{LighterRed}-0.23 (0.4) &  -0.07 (0.6) \\
   \multicolumn{3}{r|}{\textbf{Mean}}  & \cellcolor{LightGreen}\textbf{0.05} & -0.292 & \cellcolor{LighterGreen}-0.258 & -0.404 & -0.332 & \cellcolor{LightRed}-4,56 & \cellcolor{LighterRed}-0.712 & -0.526 \\

\bottomrule

  & 24  & 1  & \cellcolor{LightGreen}\textbf{0.0 (0.1)} & -0.46 (1.9)& -0.41 (1.8)& -0.12 (1.1)& -0.44 (1.9)& \cellcolor{LightRed}-2.7 (3.9)& \cellcolor{LighterRed}-0.81 (2.4)& \cellcolor{LighterGreen}-0.09 (1.0)\\
  
  & 168  & 1  & \cellcolor{LightGreen}\textbf{0.01 (0.1)} & -0.41 (1.8)& -0.4 (1.8)& -0.18 (1.2)& -0.66 (2.3)& \cellcolor{LightRed}-2.82 (3.9)& \cellcolor{LighterRed}-1.03 (2.8)& \cellcolor{LighterGreen}-0.12 (1.1)\\
  
  & 168  & 24  & \cellcolor{LightGreen}\textbf{0.01 (0.1)} & \cellcolor{LighterGreen}-0.31 (1.5)& -0.35 (1.6)& -0.61 (2.1)& \cellcolor{LighterRed}-1.28 (3.0)& \cellcolor{LightRed}-9.22 (2.5)& -0.92 (2.6)& -0.99 (2.6)\\
  
  & 744  & 1  & \cellcolor{LightGreen}\textbf{-0.08 (0.9)} & -0.38 (1.7)& -0.39 (1.8)& \cellcolor{LighterGreen}-0.25 (1.3)& -0.62 (2.2)& \cellcolor{LightRed}-3.57 (4.2)& \cellcolor{LighterRed}-1.17 (3.0)& -0.29 (1.6)\\
  
\parbox[t]{2mm}{\multirow{-5}{*}{\rotatebox[origin=c]{90}{\textbf{IP addr.}}}}   & 744  & 168  & \cellcolor{LightGreen}\textbf{-0.09 (0.9)} & -0.5 (1.8)& \cellcolor{LighterGreen}-0.34 (1.5)& \cellcolor{LighterRed}-3.97 (4.2)& -1.52 (3.2)& \cellcolor{LightRed}-9.17 (2.5)& -0.52 (1.9)& -2.77 (3.8)\\
 
\bottomrule

 %\multicolumn{3}{r|}{\textbf{Overall}}  & \cellcolor{LightGreen}\textbf{0.0 (0.1)} &  -0.09 (0.3) &  \cellcolor{LighterGreen}-0.08 (0.3) &  \cellcolor{LighterRed}-0.19 (0.4) &  -0.17 (0.4) &  \cellcolor{LightRed}-0.72 (0.4) &  -0.18 (0.4) &  -0.16 (0.4) \\
   \multicolumn{3}{r|}{\textbf{Mean}}  & \cellcolor{LightGreen}\textbf{-0.03} & -0.412 & \cellcolor{LighterGreen}-0.378 & \cellcolor{LighterRed}-1.026 & -0.904 & \cellcolor{LightRed}-5.496 & -0.89 & -0.852 \\

\bottomrule
   \multicolumn{3}{r|}{\textbf{Overall mean}}  & \cellcolor{LightGreen}\textbf{0.0173} & -0.2887 & \cellcolor{LighterGreen}-0.2573 & -0.5427 & -0.4513 & \cellcolor{LightRed}-4.808 & \cellcolor{LighterRed}-0.7313 & -0.562 \\

\end{tabular}
\end{table*}

\subsection{Evaluation of models performance}

We trained 8 models (baseline mean model and 7 DL models) across five combinations of training and prediction windows, for 18 time series metrics within 1,831 time series (283 time series for institutions, 548 time series for institutional subnets, and 1000 time series for IP addresses), resulting in 1,318,320 final models. In this section, we present the results for the \textit{Number of bytes} time series metrics (transmitter data), as it is the most commonly used metric for network traffic forecasting in both network management and anomaly detection.

%Tables~\ref{tab:RMSE} and \ref{tab:R2-SCORE} present aggregated results for RMSE and \(R^2\)-score. We define one experimental setting as a combination of the dataset part, training window, and prediction window. The reported values are the mean and standard deviation over all corresponding time series for each experimental setting. In these tables, the best-performing model is highlighted with a bold font and a green-shaded cell, while the worst-performing model is marked with a red-shaded cell. Lighter shades indicate second best or worst performance, providing a relative sense of each model's accuracy.

Tables~\ref{tab:RMSE} and \ref{tab:R2-SCORE} present aggregated results for RMSE and \(R^2\)-score respectively. Each row represents aggregated results (mean and standard deviation) of one experimental setting with specific training and prediction window combination on all time series within dataset time series type (e.g., Institutions, Institutions subnets, and single IPs time series). The best-performing model is highlighted with a bold font and green-shaded cell, while the worst-performing model is marked with a red-shaded cell. Lighter shades indicate second best or worst performance, providing a relative sense of each model's accuracy. The same values may sometimes result in multiple same shaded cells in a row.

Different evaluation metrics capture distinct aspects of forecasting quality. The resulting RMSE indicates that models such as GRU, LSTM, GRU-FCN, and LSTM-FCN perform similarly well with very close results in all scenarios. GRU and LSTM models slightly outperform their FCN-enhanced counterparts in \(R^2\)-score better-matching the trendiness of the data.

It can be seen that the results of the model performances depend on the window settings. For example, settings with a training window equal to or less than 168 values suggest that GRU-FCN and LSTM-FCN are quite effective in capturing overall trends in the data. Contrarily, the setting with a training window larger than 168 values suggests the opposite statement. The baseline Mean model excels in terms of \(R^2\)-score by accurately reflecting the general trend, although it lacks the capacity to model seasonal or cyclic variations. Among the deep learning models, RCLSTM demonstrates superior performance for one-step-ahead predictions on \(R^2\)-score. Finally, no single model consistently outperforms the others when predicting over longer horizons.

\begin{table*}
\centering
\caption{Resulting mean and standard deviation of \textit{Harmonic score} in the format ``\(<\)mean\(>\). }\label{tab:harmonic-score}
\begin{tabular}{r|r|r|C{1.3cm}|C{1.3cm}|C{1.3cm}|C{1.3cm}|C{1.3cm}|C{1.3cm}|C{1.3cm}|C{1.3cm}}
\toprule
 & \rotatebox{90}{\parbox{1.4cm}{ \textbf{Training \ window}}} & \rotatebox{90}{\parbox{1.4cm}{\textbf{Prediction \ window}}}  & \rotatebox{90}{\textbf{Mean}}  & \rotatebox{90}{\textbf{GRU}}  & \rotatebox{90}{\textbf{LSTM}}  & \rotatebox{90}{\textbf{GRU-FCN}}  & \rotatebox{90}{\textbf{LSTM-FCN}}  & \rotatebox{90}{\textbf{InceptionTime}}  & \rotatebox{90}{\textbf{ResNet}}  & \rotatebox{90}{\textbf{RCLSTM}}  \\
\toprule
  & 24  & 1  & \cellcolor{LightRed}0.14638  & 0.11988  & 0.12150  & \cellcolor{LightGreen}\textbf{0.11476}  & \cellcolor{LighterGreen}0.11606  & \cellcolor{LighterRed}0.13448  & 0.12220  & 0.11924 \\
  
  & 168  & 1  & 0.15170  & 0.12234  & 0.12238  & \cellcolor{LightGreen}\textbf{0.11722}  & \cellcolor{LighterGreen}0.11811  & \cellcolor{LighterRed}0.15712  & \cellcolor{LightRed}0.16212  & 0.11846 \\
  
  & 168  & 24  & 0.15286  & 0.15592  & 0.15702  & \cellcolor{LightGreen}\textbf{0.14076}  & \cellcolor{LighterGreen}0.14397  & \cellcolor{LightRed}0.85087  & \cellcolor{LighterRed}0.17161  & 0.16150 \\
  
  & 744  & 1  & 0.15491  & 0.12492  & \cellcolor{LighterGreen}0.12375  & 0.13182  & 0.13322  & \cellcolor{LighterRed}0.18183  & \cellcolor{LightRed}0.20099  & \cellcolor{LightGreen}\textbf{0.12041} \\
  
\parbox[t]{2mm}{\multirow{-5}{*}{\rotatebox[origin=c]{90}{\textbf{Inst.}}}}   & 744  & 168  & \cellcolor{LightGreen}\textbf{0.15642}  & 0.17122  & \cellcolor{LighterGreen}0.17116  & 0.17118  & 0.16935  & \cellcolor{LightRed}0.85523  & 0.18619  & \cellcolor{LighterRed}0.22440 \\

\bottomrule

 \multicolumn{3}{r|}{\textbf{Mean}}  & 0.15246 & 0.13885 &  0.13916 &  \cellcolor{LightGreen}\textbf{0.13515} &  \cellcolor{LighterGreen}0.13596 &  \cellcolor{LightRed}0.43591 &  \cellcolor{LighterRed}0.16862 &  0.14880 \\
 
\bottomrule
  & 24  & 1  & \cellcolor{LighterRed}0.17652  & 0.15198  & 0.15372  & \cellcolor{LightGreen}\textbf{0.14844}  & \cellcolor{LighterGreen}0.14955  & 0.17427  & 0.15523  & \cellcolor{LightRed}0.18015 \\
  
  & 168  & 1  & \cellcolor{LighterRed}0.18660  & 0.15400  & 0.15485  & \cellcolor{LightGreen}\textbf{0.15295}  & \cellcolor{LighterGreen}0.15374  & \cellcolor{LightRed}0.19104  & 0.18327  & 0.16024 \\
  
  & 168  & 24  & 0.18937  & 0.18747  & 0.18793  & \cellcolor{LightGreen}\textbf{0.17265}  & \cellcolor{LighterGreen}0.17702  & \cellcolor{LightRed}0.86831  & 0.20200  & \cellcolor{LighterRed}0.21899 \\
  
  & 744  & 1  & 0.19685  & 0.15636  & \cellcolor{LighterGreen}0.15621  & 0.16950  & 0.17137  & \cellcolor{LighterRed}0.21963  & \cellcolor{LightRed}0.22311  & \cellcolor{LightGreen}\textbf{0.15169} \\
  
\parbox[t]{2mm}{\multirow{-5}{*}{\rotatebox[origin=c]{90}{\textbf{Inst. subnets}}}}   & 744  & 168  & \cellcolor{LighterGreen}0.20168  & 0.20279  & \cellcolor{LightGreen}\textbf{0.20110}  & 0.20976  & 0.20196  & \cellcolor{LightRed}0.87231  & 0.22229  & \cellcolor{LighterRed}0.27010 \\

\bottomrule

 \multicolumn{3}{r|}{\textbf{Mean}}  &  0.19021 & \cellcolor{LighterGreen}0.17052 &  0.17076 &  0.17066 &  \cellcolor{LightGreen}\textbf{0.17049} &  \cellcolor{LightRed}0.46511 &  \cellcolor{LighterRed}0.19718 &  0.19623 \\

\bottomrule
  & 24  & 1  & \cellcolor{LightRed}0.29957  & \cellcolor{LightGreen}\textbf{0.12944}  & 0.13082  & \cellcolor{LighterGreen}0.13074  & 0.13197  & 0.15928  & 0.13441  & \cellcolor{LighterRed}0.16410 \\
  
  & 168  & 1  & \cellcolor{LightRed}0.29675  & \cellcolor{LightGreen}\textbf{0.13030}  & \cellcolor{LighterGreen}0.13137  & 0.13296  & 0.13481  & \cellcolor{LighterRed}0.16474  & 0.14476  & 0.16158 \\
  
  & 168  & 24  &\cellcolor{LighterRed} 0.29812  & 0.13804  & \cellcolor{LighterGreen}0.13796  & \cellcolor{LightGreen}\textbf{0.13748}  & 0.14576  & \cellcolor{LightRed}0.93780  & 0.14757  & 0.17450 \\
  
  & 744  & 1  & \cellcolor{LightRed}0.30066  & \cellcolor{LightGreen}\textbf{0.13010}  & \cellcolor{LighterGreen}0.13143  & 0.13599  & 0.13734  & \cellcolor{LighterRed}0.19576  & 0.15526  & 0.13802 \\
  
\parbox[t]{2mm}{\multirow{-5}{*}{\rotatebox[origin=c]{90}{\textbf{IP addr.}}}}   & 744  & 168  & \cellcolor{LighterRed}0.30275  & \cellcolor{LighterGreen}0.15091  & \cellcolor{LightGreen}\textbf{0.15003}  & 0.22075  & 0.16970  & \cellcolor{LightRed}0.94300  & 0.15623  & 0.19307 \\

\bottomrule

\multicolumn{3}{r|}{\textbf{Mean}}  & \cellcolor{LighterRed}0.29957 & \cellcolor{LightGreen}\textbf{0.13576} &  \cellcolor{LighterGreen}0.13632 &  0.15154 &  0.14385 &  \cellcolor{LightRed}0.48002 &  0.14765 &  0.16626 \\

\bottomrule
   \multicolumn{3}{r|}{\textbf{Overall mean}}  & \cellcolor{LighterRed}0.21408 & \cellcolor{LightGreen}\textbf{0.14838} & \cellcolor{LighterGreen}0.14875 & 0.15245 & 0.1501 & \cellcolor{LightRed}0.46035 & 0.17115 & 0.17043 \\

\end{tabular}
\end{table*}

%Table~\ref{tab:harmonic-score} summarizes the forecasting performance using the Harmonic-score, a metric that combines multiple evaluation aspects into a single measure. The Harmonic score results underscore the superior performance of the GRU-FCN model in most experimental settings with a training period equal to or less than 168  while also showing that LSTM-FCN, LSTM, and GRU architectures achieve very similar results in several settings. Nevertheless, the GRU-FCN significantly drops the measure compared to these models in setting with IP address dataset part, training window equal to 744, and prediction window equal to 168. This drop in the score caused GRU-FCN not to outperform other DL models in the overall mean.

Table~\ref{tab:harmonic-score} summarizes the forecasting performance using the Harmonic-score, a metric that combines multiple evaluation aspects into a single measure. The results show that the LSTM-FCN, LSTM, and GRU architectures produce very similar results in several configurations. Despite the GRU-FCN model demonstrating superior performance in most experimental settings with a training period of 168 or less, it shows a significant drop in the score in the setting involving the IP address dataset, with a training window of 744 and a prediction window of 168. This decrease prevented GRU-FCN from outperforming other DL models in the overall mean.

Overall, classical architectures such as GRU, LSTM, and their convolutionally enhanced variants achieve similarly accurate results in terms of RMSE, \(R^2\)-score, and Harmonic score. On the macro average, GRU-FCN performs best for the Institution dataset, LSTM-FCN for the Institution subnets dataset, and GRU for the IP address dataset. Considering the overall macro average across all datasets, GRU produces the most precise forecasts, followed by LSTM.

We also observe that InceptionTime is the weakest performer across all evaluation measures, followed by ResNet. InceptionTime, originally designed for time series classification, struggles with network traffic forecasting due to its reliance on multi-scale convolutions, which increase model complexity without yielding significant performance gains. Its limited ability to capture long-term dependencies further hampers its accuracy. Similarly, ResNet, despite the use of residual connections, appears less effective in modeling sequential patterns, potentially leading to suboptimal generalization at high computational costs. These findings highlight the need for architectures better suited to network time series forecasting, such as those incorporating recurrent or attention-based mechanisms.

A clear performance trend emerges across the different dataset parts, reflecting the models’ ability to capture data trendiness based on the \(R^2\)-score. Forecasting performance is highest for the Institution dataset, declines for Institution subnets, and is lowest for the IP address sample. This degradation likely stems from the increasing proportion of missing values, which are minimal in the Institution dataset but significantly more prevalent in the IP address sample. The impact of missing data is further supported by the strong negative correlation (–0.69 for the IP address dataset) between the ratio of missing samples and the \(R^2\)-score.

The choice of training and prediction windows significantly affects model performance. Larger prediction windows generally degrade forecast accuracy, as models struggle to capture long-term dependencies with the same precision as short-term predictions.

Analyzing the most and least predictable time series reveals that less predictable series often display sudden, abnormal spikes in values, while the most predictable series follow more regular patterns. This observation is supported by autocorrelation and partial autocorrelation analyses, which show that highly predictable time series typically exhibit strong seasonal behavior.

\begin{figure}
    \centering
    \includegraphics[width=0.6\linewidth]{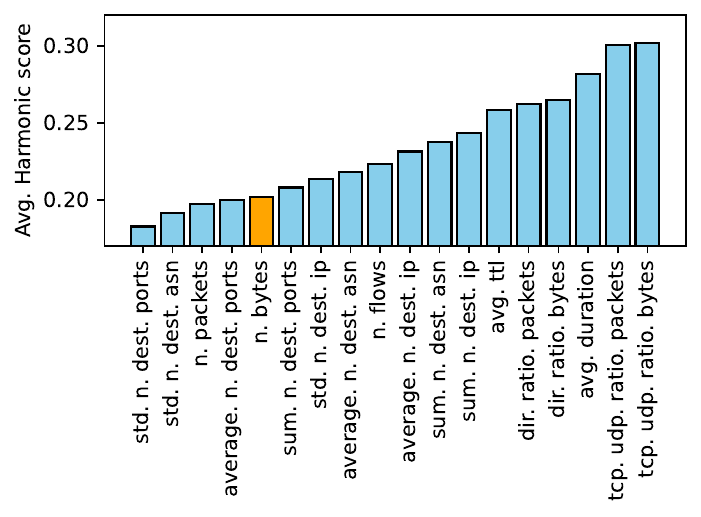}
    \caption{Average across all models of overall mean \textit{Harmonic score} for each time series metric}
    \label{fig:ts-metric-overall-comparision}
\end{figure}

While the \textit{Number of bytes} (i.e., transmitted data) is often used as a primary metric for network traffic forecasting, our results reveal that forecasting evaluation measures vary considerably across different metrics as shown in Figure~\ref{fig:ts-metric-overall-comparision}. Some metrics follow similar predictability patterns to the \textit{Nmber of bytes}, while others present additional challenges due to their inherent variability and dependence on specific network behaviors. Complete forecasting results for all metrics, along with code and supplementary analyses, are available on GitHub\footref{footnote:github}.

Metrics representing aggregate traffic volumes, such as the \textit{Number of flows} and \textit{Number of packets}, generally follow similar patterns to the \textit{Number of bytes} and are effectively captured by DL models. These metrics often exhibit seasonal behavior, making them easier to predict, especially with recurrent architectures combined with convolutional layers.

In contrast, forecasting more dynamic and structurally complex metrics, such as \textit{TCP/UDP ratios} and \textit{Direction ratios}, proves significantly more challenging. These metrics are prone to sudden fluctuations due to external influences, such as changes in user behavior. Our results show that DL models struggle to capture these fluctuations, leading to lower \( R^2 \)-score and increased forecasting error.

\subsection{Deployability}

Model deployability is a key factor; therefore, we measured the training and inference time of all DL models. The training was conducted on MetaCentrum\footnote{https://metavo.metacentrum.cz/en/index.html}, which manages the National Grid Infrastructure (NGI) as part of the European Grid Infrastructure (EGI). To ensure consistency, all experiments were run on an identical pool of machines with fixed configurations. Each run used a dedicated virtual environment with 4 exclusive CPUs, minimizing interference from other processes and ensuring comparable results.

Table~\ref{tab:Deploymentability} compares the relative training and prediction times of each model, normalized to 100 data points over all experiments. These metrics highlight the computational feasibility of each model for real-world deployment. Models with long training times may be impractical for frequent retraining, while high prediction latency can hinder responsiveness, particularly when forecasting a single next value. The results show that GRU-FCN consistently achieves the shortest training and prediction times across all dataset segments and experiments, making it a strong candidate for scenarios with limited time or computational resources.

\begin{table}
\centering
\caption{Relative training and prediction times in seconds (normalized per 100  datapoints)}\label{tab:Deploymentability}
\begin{tabular}{r|c|c|c|c|c|c}
\toprule
  & \multicolumn{3}{c|}{\textbf{ Training }} & \multicolumn{3}{c}{\textbf{ Prediction }}  \\
  & \textbf{Inst.} & \textbf{Sub.} & \textbf{IP} &  \textbf{Inst.} & \textbf{Sub.} & \textbf{IP}  \\
\toprule

\textbf{GRU} & 4.249  & 4.509  & 4.27  & 0.113  & 0.128  & 0.147 \\
 \textbf{LSTM} & 5.327  & 4.276  & 4.507  & 0.109  & 0.116  & 0.107 \\
 \textbf{GRU-FCN} & \cellcolor{LightGreen}\textbf{1.196}  & \cellcolor{LightGreen}\textbf{0.958}  & \cellcolor{LightGreen}\textbf{1.046}  & \cellcolor{LightGreen}\textbf{0.065}  & \cellcolor{LightGreen}\textbf{0.052}  & \cellcolor{LightGreen}\textbf{0.054} \\
 \textbf{LSTM-FCN} & 3.188  & 3.376  & 3.127  & 0.122  & 0.134  & 0.121 \\
 \textbf{InceptionTime} & \cellcolor{LighterGreen}2.936  & \cellcolor{LighterGreen}3.015  & \cellcolor{LighterGreen}2.957  & \cellcolor{LighterGreen}0.075  & \cellcolor{LighterGreen}0.079  & \cellcolor{LighterGreen}0.084 \\
 \textbf{ResNet} & \cellcolor{LighterRed}6.369  & \cellcolor{LighterRed}5.956  & \cellcolor{LighterRed}5.493  & \cellcolor{LightRed}0.318  & \cellcolor{LightRed}0.310  & \cellcolor{LightRed}0.328 \\
 \textbf{RCLSTM} & \cellcolor{LightRed}11.371  & \cellcolor{LightRed}8.448  & \cellcolor{LightRed}16.287  & \cellcolor{LighterRed}0.303  & \cellcolor{LighterRed}0.288  & \cellcolor{LighterRed}0.286 \\

\bottomrule

\end{tabular}
\end{table}

\section{Discussion} \label{sec:findings}

In this section, we discuss key findings related to the four research questions (RQs) investigated in this study. We examine factors affecting forecasting precision, compare the performance of state-of-the-art deep learning models on a public real-world network dataset, and highlight the practical implications of our results for network traffic forecasting.

\textbf{RQ1:} \textit{What is the performance of state-of-the-art deep learning models for real-world network traffic forecasting?}

Our evaluation shows that state-of-the-art deep learning models can achieve competitive forecasting accuracy for real-world network traffic. However, no single model consistently outperforms others across all experimental settings, including dataset partitions, training windows, and prediction windows. The results suggest that while deep learning models can provide accurate forecasts, their effectiveness depends heavily on the specific experimental conditions.

GRU and LSTM architectures achieve the best overall results, while their hybrid variants with fully convolutional networks (GRU-FCN and LSTM-FCN) perform slightly better on the Institution and Institution subnets timeseries. Notably, GRU-FCN offers a balanced trade-off between accuracy and computational efficiency. In contrast, models like InceptionTime and ResNet underperform in most settings, possibly because their hierarchical feature extraction mechanisms are less suited for capturing the temporal dependencies and variability in network traffic data.

\textbf{RQ2:} \textit{How do the training and prediction windows impact the forecasting performance of state-of-the-art deep learning models?}

The selection of training and prediction windows significantly impacts model accuracy, balancing the trade-off between capturing long-term dependencies and managing computational overhead. Short training windows often fail to capture temporal patterns beyond immediate fluctuations, leading to suboptimal performance. Conversely, excessively long training windows introduce noise and increase computational costs without substantial accuracy gains. However, our results show that the highest precision was achieved with small training windows, while increasing the window size led to a decline in performance.

For prediction windows, we observe a consistent trend: forecasting single next value yields significantly higher accuracy than longer prediction horizons. The degradation in accuracy over extended prediction windows is primarily due to error accumulation, a phenomenon observed across all evaluated DL models. Notably, LSTM and GRU exhibit the most stable performance across different prediction windows, highlighting their robustness in handling extended forecasts.

Our findings suggest that much of the network traffic time series lacks strong seasonality or contains large outliers, which models with short training and prediction windows handle more effectively. As a result, larger windows perform worse in this study. However, longer training and prediction windows could be advantageous in real-world deployments, as they require fewer inferences and are less reactive to abnormal spikes, making them potentially useful for anomaly detection.

\textbf{RQ3:} \textit{How does the granularity of aggregation (e.g., individual IP addresses, subnets, or institutions) impact the forecasting performance of state-of-the-art deep learning models?}

The forecasting precision of DL models is strongly influenced by the level of aggregation. Our results show that the ability to capture trend patterns is highest at the institution level, decreases at the subnet level, and is lowest at the individual IP address level.

This trend is primarily due to increasing data variability and sparsity at finer granularities. Institution-level traffic patterns are generally more stable and exhibit clear periodic trends, enabling higher predictive accuracy. In contrast, individual IP addresses often contain irregular spikes and a higher proportion of missing values, making them significantly harder to forecast. This aligns with our correlation analysis, which shows a strong negative relationship ($-0.69$) between missing data and \( R^2 \)-score at the IP address level. Models trained on IP-level data must contend with high noise levels and unpredictability, leading to increased forecasting errors.

These findings suggest that different modeling strategies may be required for different hierarchical levels. Institution-level forecasting may be well-suited to simpler recurrent models, while IP-level forecasting may require more advanced architectures or preprocessing techniques to handle sparsity and irregularities effectively.

\textbf{RQ4:} \textit{How do different monitored metrics (e.g., transmitted data, TCP/UDP ratio, average duration) affect the forecasting performance of state-of-the-art deep learning models?}

Forecasting accuracy varies significantly across different network traffic metrics. Metrics with clear seasonal patterns, such as the number of flows, packets, and bytes, are well predicted by DL models. In contrast, highly variable metrics, such as TCP/UDP ratios, pose challenges due to their irregular fluctuations and external dependencies, resulting in lower forecasting precision. Aggregation level also plays a crucial role—metrics at the institution level are more predictable than those at the IP level due to reduced noise. Similarly, longer aggregation windows improve performance for volume-based metrics but may obscure short-term variations in ratio-based metrics.

Overall, DL models perform well on structured, high-volume metrics but struggle with fine-grained or bursty characteristics. Future research should explore adaptive strategies to improve forecasting for complex metrics, particularly assessing whether independently optimized DL models per metric outperform multivariate forecasting approaches.

\section{Conclusion} \label{sec:conclusion}

In this study, we conducted a comprehensive evaluation of seven DL models for real-world network traffic forecasting. Our analysis revealed that LSTM and GRU surpass more complex architectures. Moreover, the GRU-FCN model achieved the best balance between predictive accuracy and computational efficiency. Beyond establishing a reproducible benchmark for network traffic forecasting using the real-world CESNET-TimeSeries24 dataset, our work lays the foundation for future work. Based on our findings, we mark several open challenges:

\begin{enumerate}
    \item \textit{Multi-time-series models}: Maintaining a separate model for each time series is both computationally expensive and operationally impractical. Future work should focus on developing a universal model capable of handling different instances within the same time-series aggregation without need for retraining.

    \item \textit{Univariate vs. multivariate models}: Training separate models for each time series metric allows fine-tuning for specific behaviors but lacks scalability, requiring extensive computational resources and maintenance. Future work should explore the trade-offs between univariate and multivariate models for network traffic forecasting.

    \item \textit{Handling Heterogeneous Data:} Effectively managing large-scale, heterogeneous time series with high missing value rates is crucial. While imputing gaps with zeros can degrade model performance, it remains valuable for detecting outages and other critical events.

    \item \textit{Improving Interpretability:} Deep learning models often function as black boxes, offering little insight into their decision-making process. Enhancing interpretability is crucial for resource allocation and anomaly detection, as it improves trust, facilitates debugging, and enables more informed decision-making.

\end{enumerate}

%Bibliography
\bibliographystyle{unsrt}  
\bibliography{main}

\end{document}